\documentclass[conference]{IEEEtran}

\usepackage{amsfonts}
\usepackage{graphicx,times,amsmath} 
\usepackage{latexsym}
\usepackage{amssymb}
\usepackage{color,afterpage}

\IEEEoverridecommandlockouts

\begin{document}
\title{\ \\ \LARGE\bf Automatic Generation of Level Maps with the\\ Do What's Possible Representation}
\author{Daniel Ashlock, Senior Member IEEE and Christoph Salge, Member
  IEEE \thanks{Daniel Ashlock is at the University of Guelph,
    dashlock@uoguelph.ca} \thanks{Christoph Salge is at the University
    of Hertfordshire,christophsalge@gmail.com} \thanks{The authors
    thank the Canadian Natural Sciences and Engineering Research
    Council of Canada (NSERC) and the European Commission (Grant INTERCOGAM) for supporting this work.}}

\date{}

\maketitle

\begin{abstract}
Automatic generation of level maps is a popular form of automatic
content generation.  In this study, a recently developed technique
employing the {\em do what's possible} representation is used to
create open-ended level maps.  Generation of the map can continue
indefinitely, yielding a highly scalable representation.  A parameter
study is performed to find good parameters for the evolutionary
algorithm used to locate high quality map generators.  Variations on
the technique are presented, demonstrating its versatility, and an
algorithmic variant is given that both improves performance and
changes the character of maps located.  The ability of the map to
adapt to different regions where the map is permitted to occupy space
are also tested.
\end{abstract}

\section{Introduction}
\label{Intro}

\begin{figure}
\centerline{\includegraphics[width=0.50\textwidth]{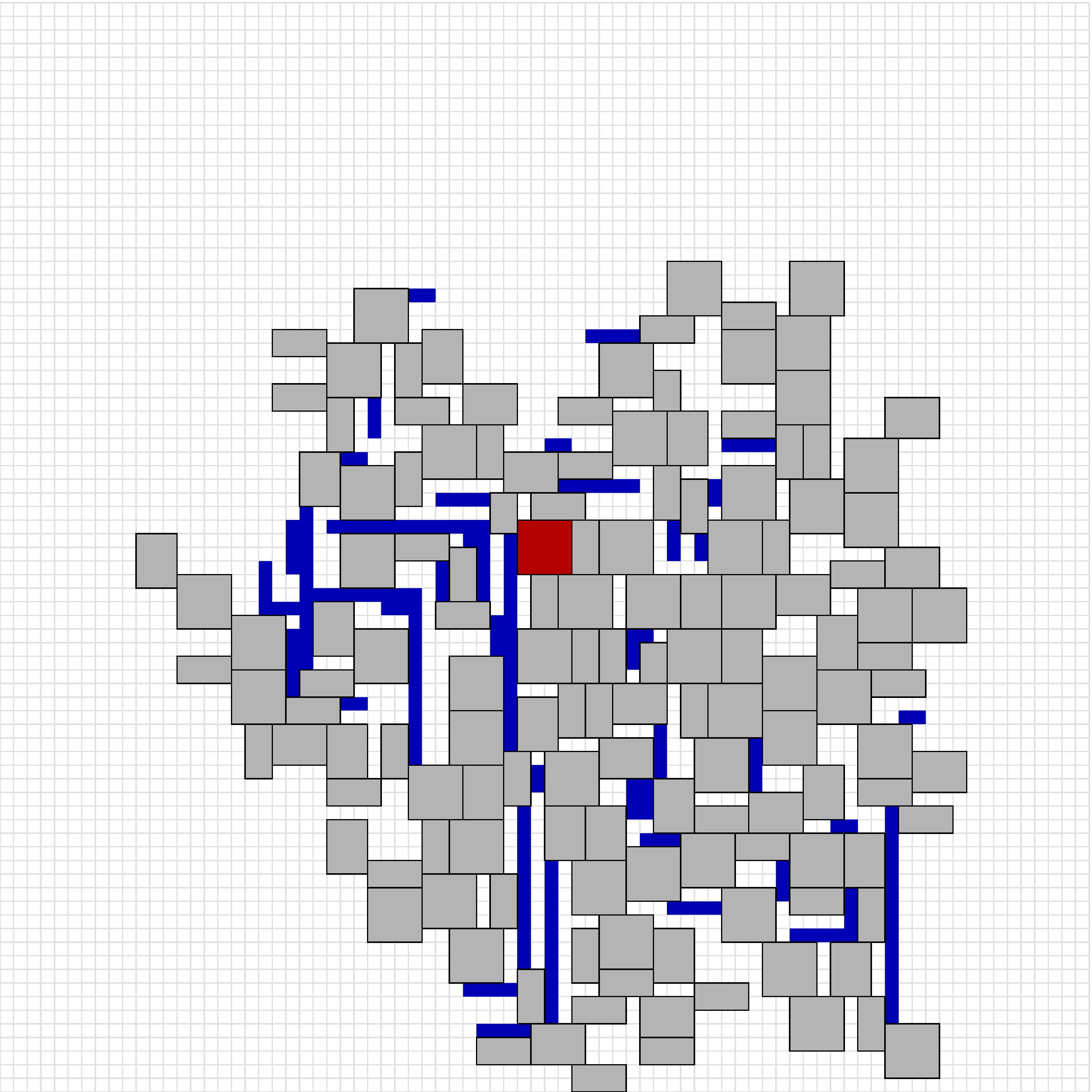}}
\caption{An example of a map evolved using the DWP representation. Red is the starting room, grey blocks are rooms and blue blocks are corridors.}
\label{Xmap}
\end{figure}

This study uses a recently developed, open-ended, generative
representation called the {\em do what's possible} (DWP)
representation \cite{AshlockDWP16} to encode level maps.  The open
ended nature of the representation encodes a sequence of maps, with
the next map in the sequence possessing one additional room.  The map
generator can be run until there are enough rooms or enough space has
been filled.  An example of one of these maps appears in Figure
\ref{Xmap}.  A strength of this representation is that it can generate
arbitrarilly large numbers of rooms by simply permitting it to run
longer.

The DWP representation has two parts.  An evolvable character
generator that will always produce a next character when queried,
called the {\em complex string generator} or CSG.  In this study the
characters are bits, zero or one, used to build integers of any
desired size to drive a serial choice process.  The CSG is partnered
with a {\em generative possibility filter} or GPF that interprets the
characters as a sequence of generative commands, rejecting the
impossible ones.  The CSG supplies pattern and the GPF filters it to
create admissible results.  While the evolutionary process used to
select the complex string generators is stochastic in the usual
fashion, the complex string generators themselves are deterministic.
This means that the maps they produce are entirely repeatable.

The map generators in this study use binary CSGs.  The bits generated
are compiled into integers of various sizes to drive a series of
decisions: which existing room will we place a new room next to, what
side of it, how is the new room situated, and what are its dimensions.
Once these decisions have been made, there is either space for the new
room, in which case it is added, or there is not, in which case the
new room is rejected.  The GPS consists of tracking occupancy of the
map and rejecting rooms that try to occupy already occupied space.
This relatively simple scheme permits the generation of huge maps
using relatively small data structures.

The remainder of this study is structured as follows.  Section
\ref{Back} reviews work in automatic level generation and earlier work
with the DWP representation.  Section \ref{XD} specifies the
experimental design including the details of the representation and
the map generation problem.  Section \ref{RandD} presents and discusses
results.  Section \ref{CNS} gives conclusions and outlines next steps.

\section{Background}
\label{Back}

Procedural content generation (PCG) consists of finding algorithmic
methods of generating content for games. It is a form of generative
design~\cite{McCormack2005GenerativeDA} often applied to game content,
and often cares about both aesthetic and functional criteria.  Search
based PCG \cite{TogeliusSBPCG} uses search methods rather than
composing algorithms that generate acceptable content in a single
pass.  Both sorts of content generation can exhibit scaling problems
which can be addressed by problem decomposition with an off-line and
an on-line phase to the software or through the use of open ended
representations, as in this study.

Automated level generation in video games can arguably be traced back
to a number of related games from the 1980s (Rogue, Hack, and
NetHack), collectively called {\em Roguelike games}.  The task is
currently of interest to the research community.  In
\cite{Sorenson10a} levels for 2D sidescroller and top-down 2D
adventure games are automatically generated using a two population
feasible/infeasible evolutionary algorithm. Answer set programming is another approach to dungeon generation~\cite{smith2014logical}. In \cite{Julian10a}
multiobjective optimization is applied to the task of search-based
procedural content generation for real time strategy maps.

This study gives an alternate representation for tasks similar to those done in
\cite{Ashlock10pcg1} which introduced checkpoint based fitness
functions for evolving maze-like levels.  This work was extended in
\cite{Ashlock10pcg2} by having multiple types of walls that defined
multiple mazes that co-existed in a single level map.

Related work includes \cite{Ashlock11pcg1} which prototyped tile
assembly to generate large maps and \cite{Ashlock13pcg} which gave a
state conditioned representation that could generate level maps of
landscape height maps.  A very different sort of representation,
based on cellular automata, is used in \cite{Johnson10} and
generalized in \cite{Ashlock15fbca, Ashlock16fbca}.

\subsection{Past work with the DWP representation}

The original publication on the DWP representation~\cite{AshlockDWP16}
used the representation on a variety of problems.  It was capable of
solving self avoiding walk problems~\cite{Ashlock06a} of unprecedented
size.  Where $12\times 12$ had been a practical limit with even adaptive
representations~\cite{Ashlock16agr}, the DWP representation solved
$40\times 40$ cases of the problem.  Applied to the problem of evolving
a Gray code~\cite{Wilson01} the DWP representation enabled the
evolution of an 8-bit Gray code with 256 members.  The DWP problem was
also used to evolve entropically rich binary strings and a type of
classification character for DNA in the initial study.  A later study
refined the systems power for DNA classification~\cite{AshlockDWP17}.

\begin{figure}[htb]
\centerline{\includegraphics[width=0.32\textwidth]{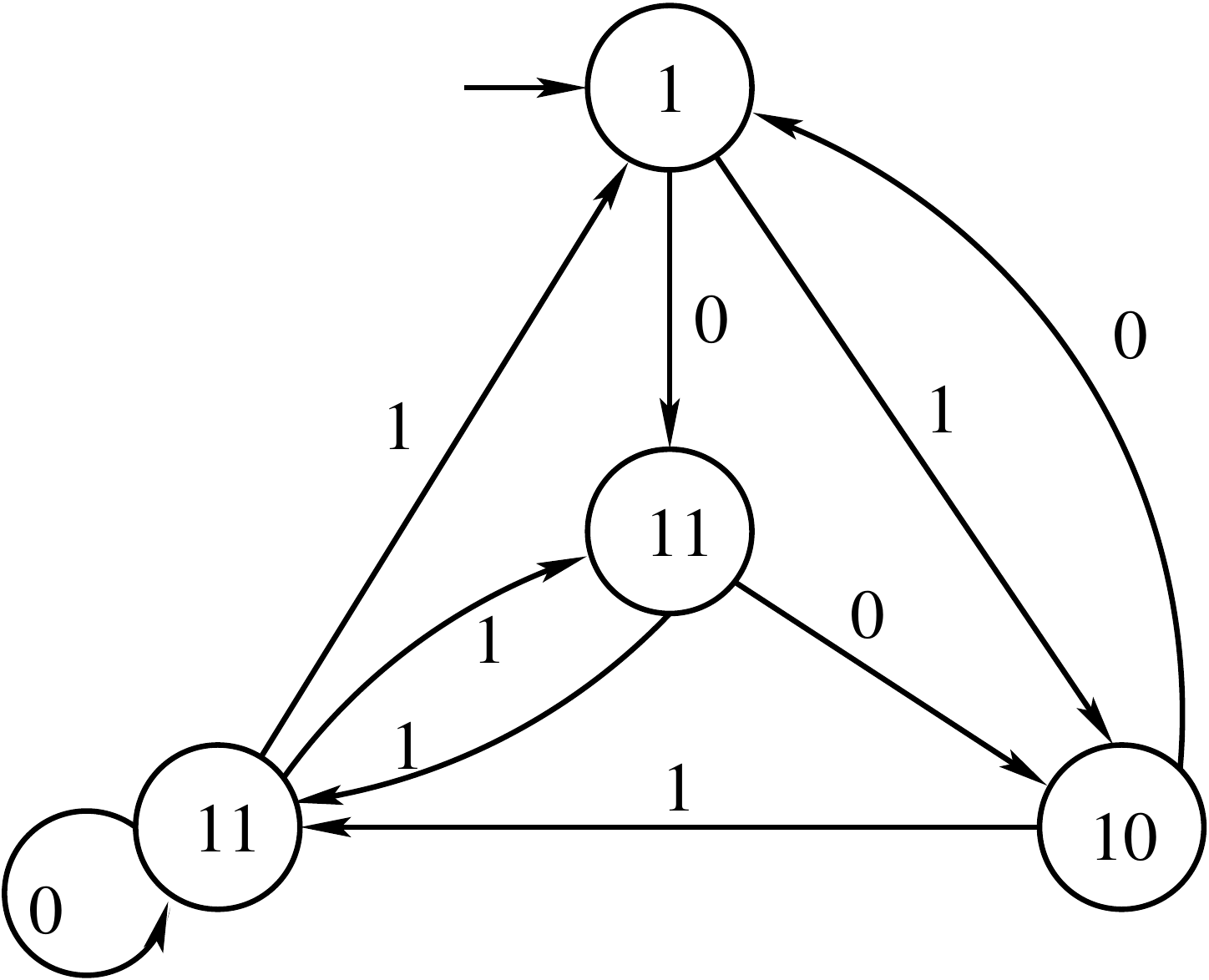}}
\centerline{\small \bf 1101111110111111...}  
\caption{An example of a self driving automaton over the binary
  alphabet and the first few characters that it emits.  The sourceless
  arrow denotes the initial state. The strings emitted are shown on
  the states.  The automata's output is used as its input.}
\label{SDA1}
\end{figure}

\section{Experimental Design}
\label{XD}

In this section we describe the representation, the problem, the
evolutionary algorithm, and the experiments performed.

\subsection{Self Driving Automata}
\label{SDA}

Figure \ref{SDA1} shows an example of a self driving automata (SDA).
This automata uses a Moore architecture, associating emitted symbols
with states.  The initial state is the topmost, pointed to by a
sourceless arrow.  When the machine starts generating bits it emits a
1. That one becomes the next input and the machine transitions to the
state labelled ``10'', emitting a 1 and a 0.  The output is compiled
and used, one bit at a time to drive transitions.  The automata
generates as many bits as required.  The emission, on some states, of
multiple bits permits the automata to encode quite complex patterns
including aperiodic ones.  The example automata in Figure \ref{SDA1},
for example, emits isolated zeros that are spaced ever farther apart
and so aer aperiodic.

The representation used for SDAs is a linear list of states, each
comprising a string to emit and the state to transition to in the
event of an input of 0 or 1.  The emitted strings associated with each
state have length 1 or 2 with that length being selected uniformly at
random, then filled with random characters.  The crossover used by
evolution on this representation is two point crossover of the list of
states, associating the initial string emitted with the first
state. Mutation consists of changing one emitted string or one
transition. The number of states an SDA is permitted to use is an
experimental parameter.  Using a linear representation for the SDA,
rather than a directed graph representation, yields crossover and
mutation operators that are much simpler to code and which permit
greater heritability of fitness.  Direct digraph representations
usually require disruptive repair operators.

\subsection{Problem Description}
\label{PS}

The problem in this study is to place the rooms and corridors of a
level map on a grid.  Rooms are rectangles with side lengths from two
to four; corridors are one wide and up to sixteen squares long.  The
process starts with an initial room placed near the center of the
grid.  The evolving part of this is a self driving automata which
generates bits that are used to make the following decisions:
\begin{enumerate}

\item An eight-bit integer modulo the current number of rooms is used
  to select the room next to which a new room is to be added.  At most
  256 rooms are permitted; the number of bits used and rooms permitted
  are easy to change.

\item A two-bit integer is used to decide if the new room will be
  above, below, left, or right of the selected room.

\item A three-bit integer is used to decide if the room is a corridor
  or not.  If all three bits are zero, the room is a corridor.

\item If the room is not a corridor, two two-bit integers are used to compute its dimensions, from $2\times 2$ to $4\times 4$.  The dimension 2 is twice as likely as 3 or 4.  If the room is a corridor a single four-bit integer is used to determine its length.  Corridors run away from the room they adjoin.

\end{enumerate}

Once a series of decisions have been made and a room proposed by the
CSG, the GPS checks to see if (i) the location of the proposed room
avoids other rooms, and (ii) is completely on the grid that is being
populated with rooms.  If the proposed room passes both these checks
then it is placed, otherwise it is rejected.

The fitness function used to judge the quality of a map in this study
is based on the total area $A$ of rooms and corridors placed on the
grid and the area $B$ {\em bounding box} of the aggregate map.  The
bounding box is the smallest rectangle on the grid that contains all
the rooms.  The fitness function is
\begin{equation}
fitness=A^2/B
\end{equation}
which encourages the placement of many rooms in a compact format.

\subsection{The Evolutionary Algorithm}

The evolutionary algorithm operates on a population of self driving
automata; varying the population size is one of the experimental
parameters.  As noted in Section \ref{SDA}, during population updating
two point crossover is used and from 1 to $MNM$ mutations are
performed on each new population member where $MNM$, the {\em maximum
  number of mutations}, is another parameter under study.  The number
of mutations is selected uniformly at random in the range one to the
maximum.

The algorithm is steady state, using mating events consisting of size
seven single tournament selection.  Seven population members are
chosen and the two best reproduce to replace the two worst.  Unless
otherwise stated, 10,000 such updatings are used.  This is a
relatively modest number of updatings which permitted the completion
of a parameter study.  Longer runs were performed after a good set of
parameters were determined.  Each experiment consists of thirty
independent runs of the evolutionary algorithm.  With two exceptions,
each fitness evaluation was initialized with a single $4\times 4$ room.

With the exception of the experiment run for additional time, all the
experiments took place on an $80\times 80$ grid with a maximum of 256
rooms.  One experiment was run for ten times as long, using a
$120\times 120$ grid with a maximum of 800 rooms and an algorithmic
modification called the recent room hack.  An experiment with standard
parameters was also performed using the recent room hack.  The recent
room hack modified the room selection process to use only four bits
to select the focal room for expansion of the map and restricted the
choice of room to the ten most recently produced rooms. This
substantially enhanced fitness for reasons discussed in Section
\ref{RandD}.

\begin{table}
\caption{Experimental parameters used.  The abbreviation RRH refers
  to the recent room hack.  Experiments 1-15 and 17-21 form the two
  parameter studies.}
\label{Xparam}
\begin{center}
\begin{tabular}{|c|c|c|c|c|}\hline
    &Pop.&     &&\\
Exp.&Size&$MNM$&States&RRH\\ \hline
 1&10&1&12&\\
 2&32&1&12&\\
 3&100&1&12&\\
 4&320&1&12&\\
 5&1000&1&12&\\
 6&10&3&12&\\
 7&32&3&12&\\
 8&100&3&12&\\
 9&320&3&12&\\
10&1000&3&12&\\
11&10&5&12&\\
12&32&5&12&\\
13&100&5&12&\\
14&320&5&12&\\
15&1000&5&12&\\ \hline
16&32&1&12&$\surd$\\ \hline
17&32&1&4&$\surd$\\
18&32&1&8&$\surd$\\
19&32&1&12&$\surd$\\
20&32&1&16&$\surd$\\
21&32&1&20&$\surd$\\ \hline
22&32&1&16&$\surd$*\\ \hline
23&32&1&16&$\surd$**\\\hline
\multicolumn{5}{l}{*Extended run}\\
\multicolumn{5}{l}{**uses a nonstandard starting room}\\
\end{tabular}
\end{center}
\end{table}

\subsection{Experiments performed}

The experiments include two parameter studies, one exploring
population size and mutation rate, while the other, performed with the
best parameters from the first, investigated the number of states used in
the SDAs.  The experiment testing the random room hack was
performed after the first parameter study and it was incorporated into
the study on the number of states used.  Table \ref{Xparam} gives the
parameters used.  The last two experiments, subsequent to both
parameter setting experiments, were a run with longer evolution time,
more rooms, and the recent room hack, and a standard run, using the
recent room hack, in which a $40\times 2$ initial room replaced the
usual one.

\begin{figure*}
  \centerline{\includegraphics[width=0.40\textwidth,angle=90]{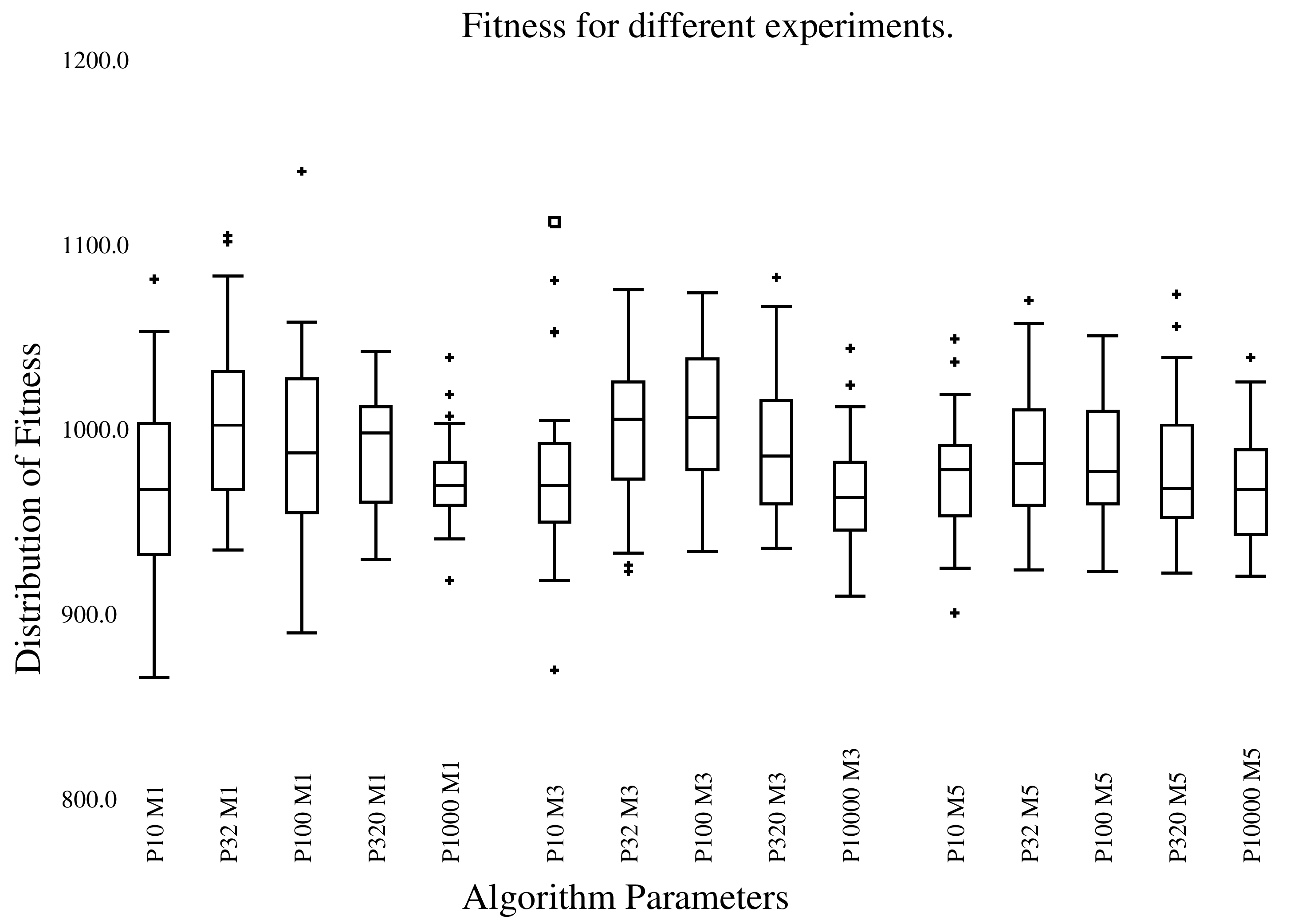}}
\caption{Results of the parameter study on population size and
  mutation rate.  Shown are box plots of the best fitness values
  from 30 runs of the evolutionary algorithm for each set of
  parameters.}
\label{Param1}
\end{figure*}

\section{Results and Discussion}
\label{RandD}

\begin{figure}[htb]
\centerline{\fbox{\includegraphics[width=0.32\textwidth]{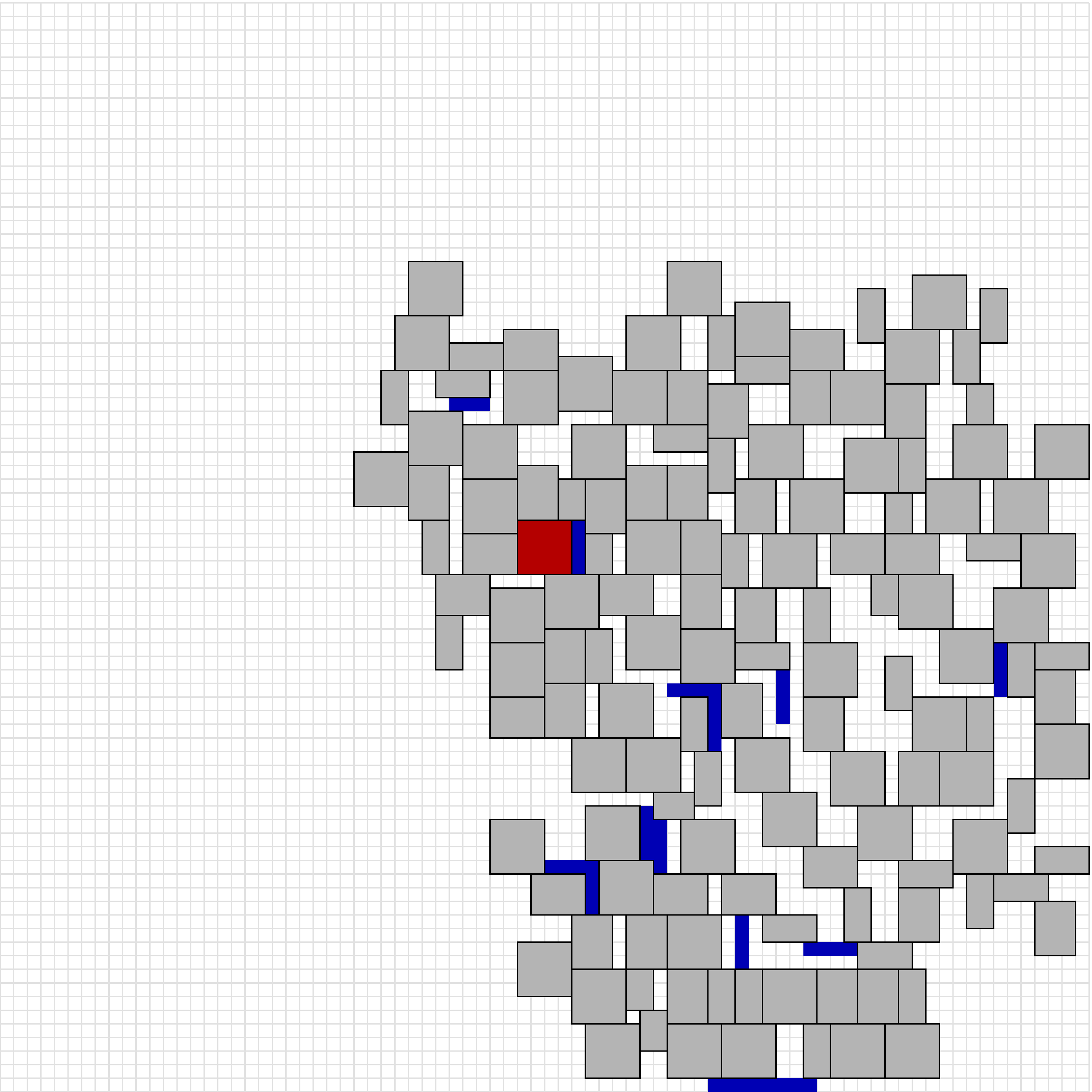}}}
\vspace{0.1in}  
\centerline{\fbox{\includegraphics[width=0.32\textwidth]{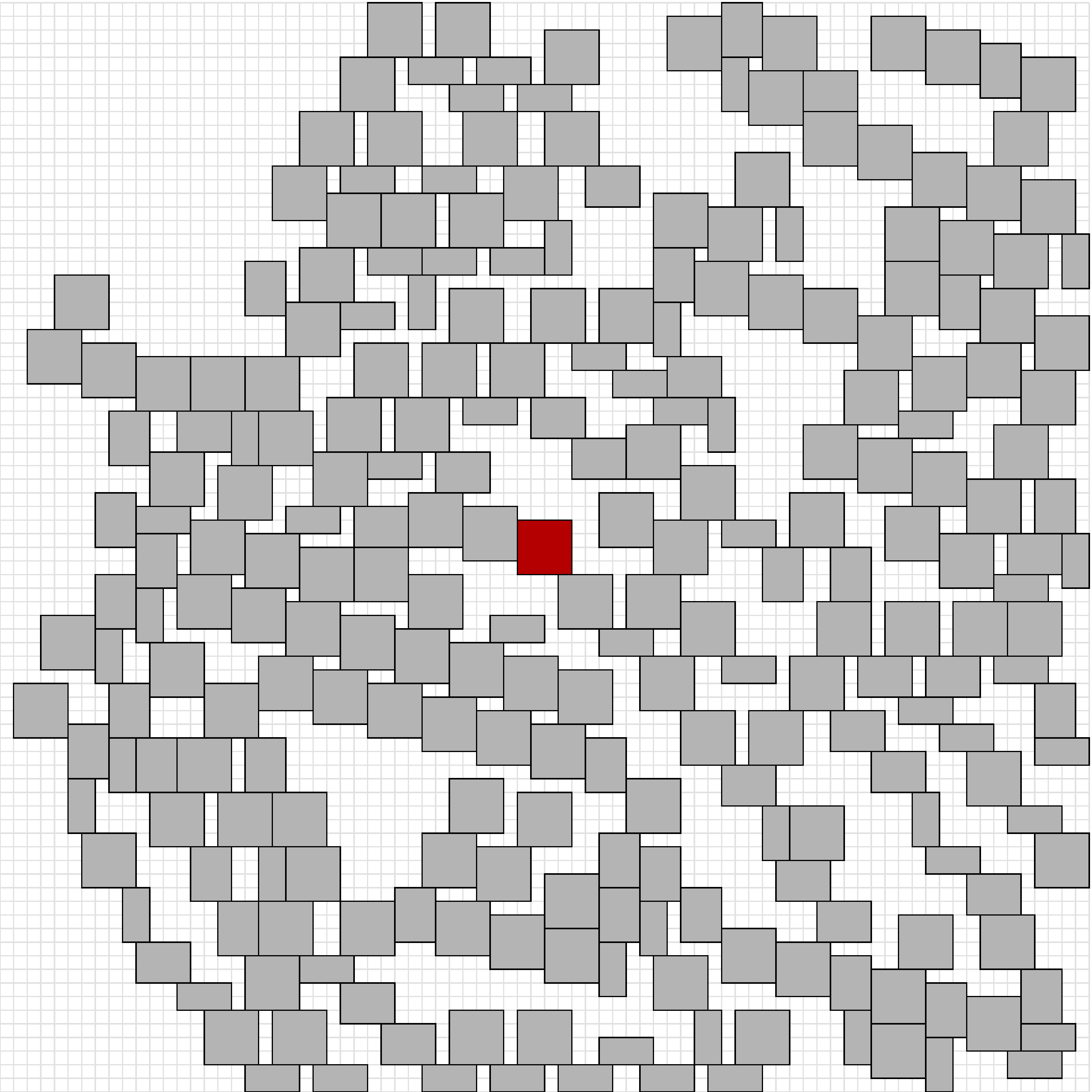}}}
\caption{Examples of maps created before the recent room hack (upper)
  and after it (lower).}
\label{BAF}
\end{figure}

The results of the first parameter study are shown in Figure
\ref{Param1}.  This study shows that the algorithm is moderately
robust to change of population size and mutation rate, with the lowest
mutation rate and the population sizes 32 and 100 being the best
performers; at the best mutation rate the population size 32 is
marginally better and so subsequent experiments used population size
32 and $MNM=1$.  A population size of 10 was clearly the worst among
the sizes tested.

\begin{figure}[htb]
  \centerline{\includegraphics[width=0.45\textwidth]{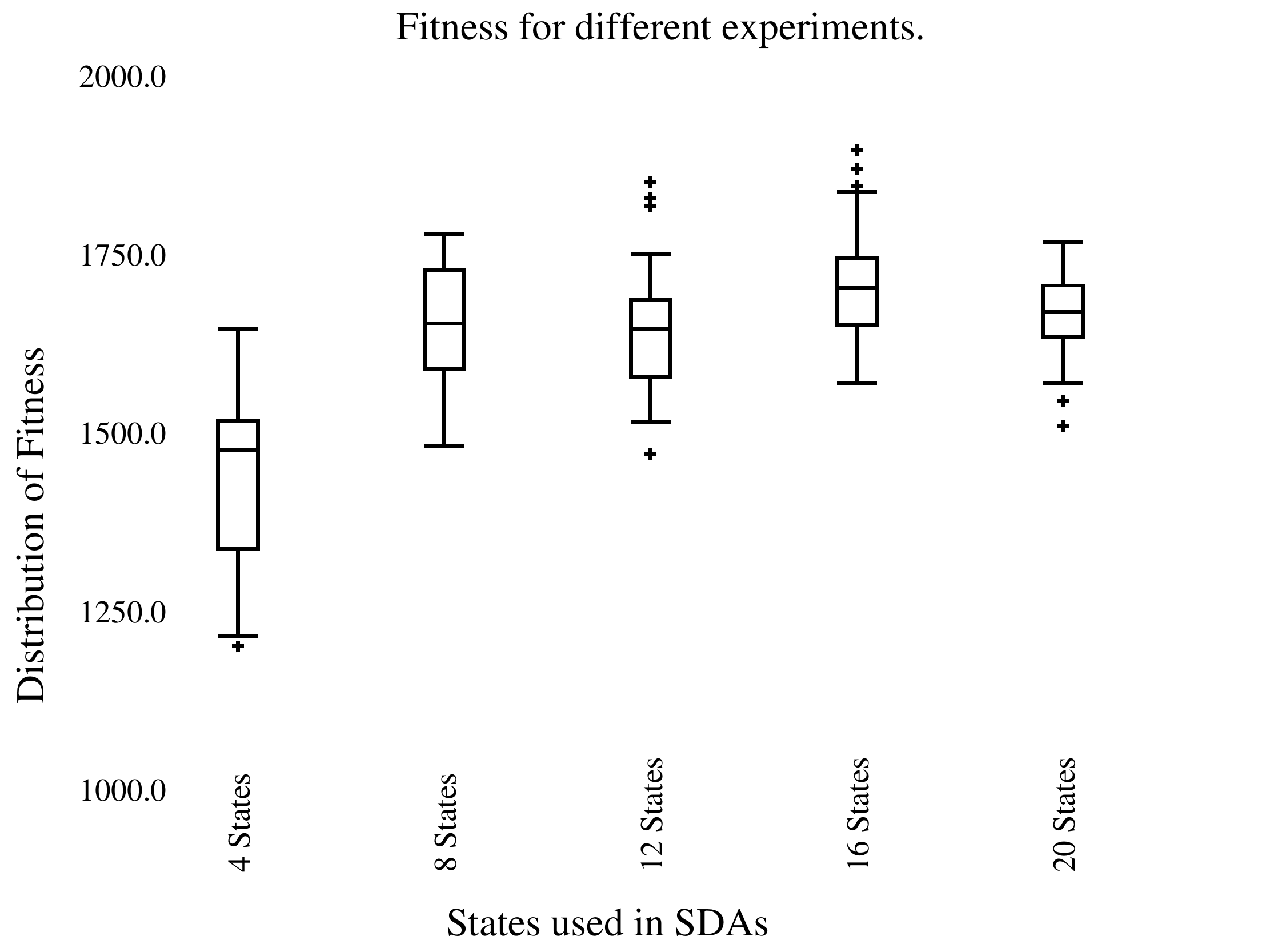}}
\caption{Results of the parameter study on number of states in the
  SDA.  Shown are box plots of the best fitness values from 30 runs of
  the evolutionary algorithm for each set of parameters.}
\label{Param2}
\end{figure}

Experiment 16, where the recent room hack was added to the algorithm,
showed about a 50\% increase in the best fitness and the number of
rooms deployed improved substantially.  Figure \ref{BAF} shows an
example of the types of maps that appear with and without the recent
room hack.  Contrasting Figure \ref{Param1} -- the parameter study
performed without the recent room hack and Figure \ref{Param2} which
depicts a parameter study with the recent room hack, notice that all
the fitness values in the first study are lower than all the fitness
values in the second study.

The recent room hack addresses a simple problem in the developmental algorithm for the maps used in this study.  As the maps fill in, many of the older room have no space next to them for the GPF will reject almost any suggestion by the CSG for a room in their area.  Permitting only recent rooms both reduces the number of bits required from the CSG to specify a room and substantially increases the change the room selected will have empty space next to it.  In retrospect, the recent room hack is an obvious improvement to the developmental algorithm for creating maps.

Figure \ref{Param2} shows the results of the parameter study examining
the best number of states to use.  It is clear that four states are
not enough, but the algorithm is otherwise fairly robust, at least at
10,000 mating events, to the number of states used with 16 states
turning in the best performance.  It may be that if the parameter
study were reperformed with additional evolutionary time, additional
states would have higher utility because it takes longer to discover
strategies that use them.

\begin{figure*}[p]
\centerline{\includegraphics[width=0.54\textwidth]{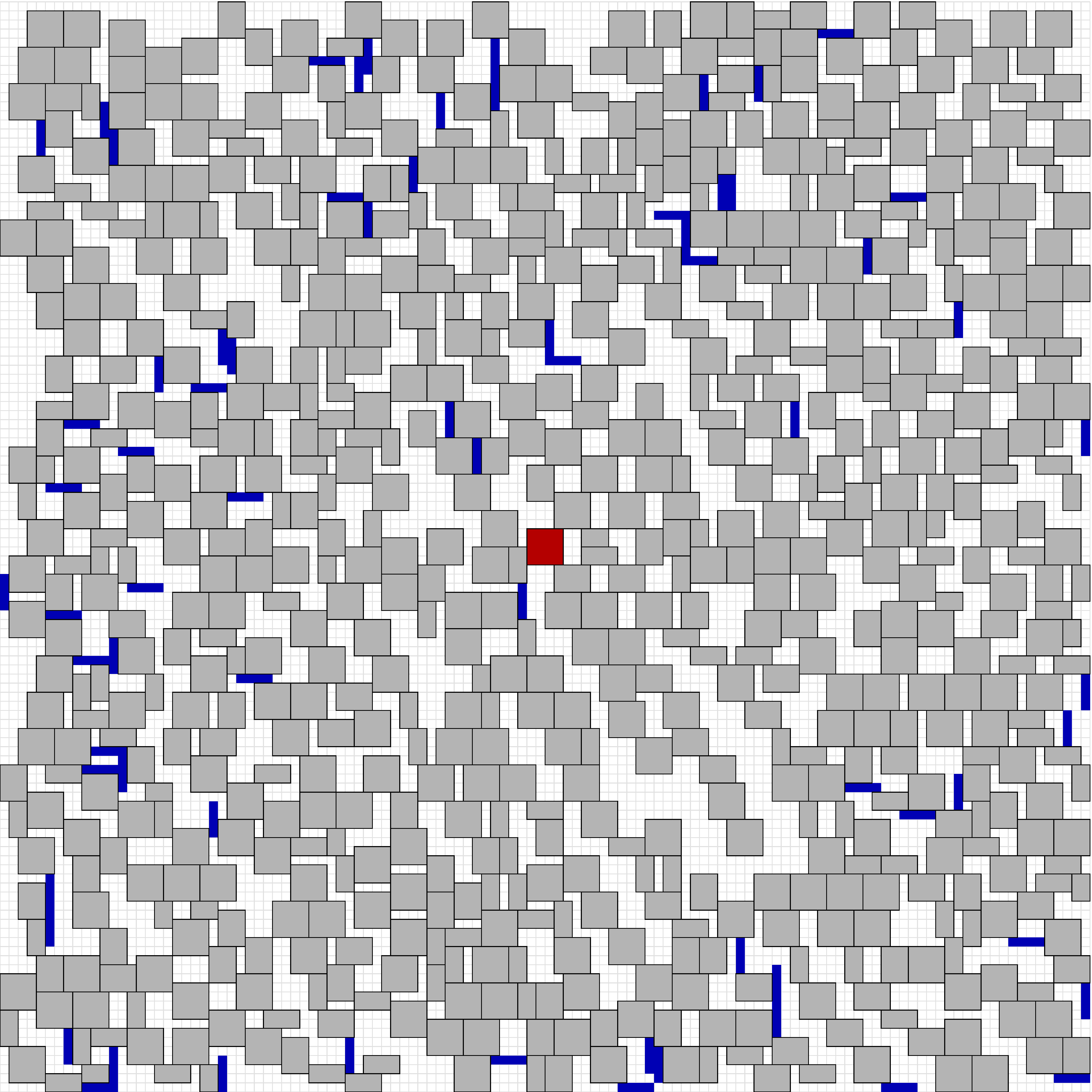}}
\caption{The map shown is rendered on a larger grid was permitted up
  to 800 rooms during evolution of the map generator.}
\label{biggie}
\end{figure*}

\subsection{Corridors, present and absent}

Looking at the examples of maps in this study, some use corridors
extensively, others do not use them at all.  Both sorts of maps
represent high-fitness local optima.  The fact that the SDAs are
deterministic and generate strings with repeated patters means that
the maps also have a relatively small number of structural elements in
each map.  If these structural elements contain corridors then the map
will as well.

\subsection{Maps made with alternate algorithm settings}

To test the notion that additional evolutionary time might help,
Experiment 22 used the recent room hack, permitted up to 800 rooms,
ran for 100,000 mating events, and used a 120$\times$120 grid instead
of an $80\times 80$ grid.  The fitness values were astronomically
higher, but also not comparable with those in the earlier experiments.
An example of a larger DWP map produced in this run is is given in
Figure \ref{biggie}.

\begin{figure*}[p]
\begin{center}
\begin{tabular}{cc}
\includegraphics[width=0.30\textwidth]{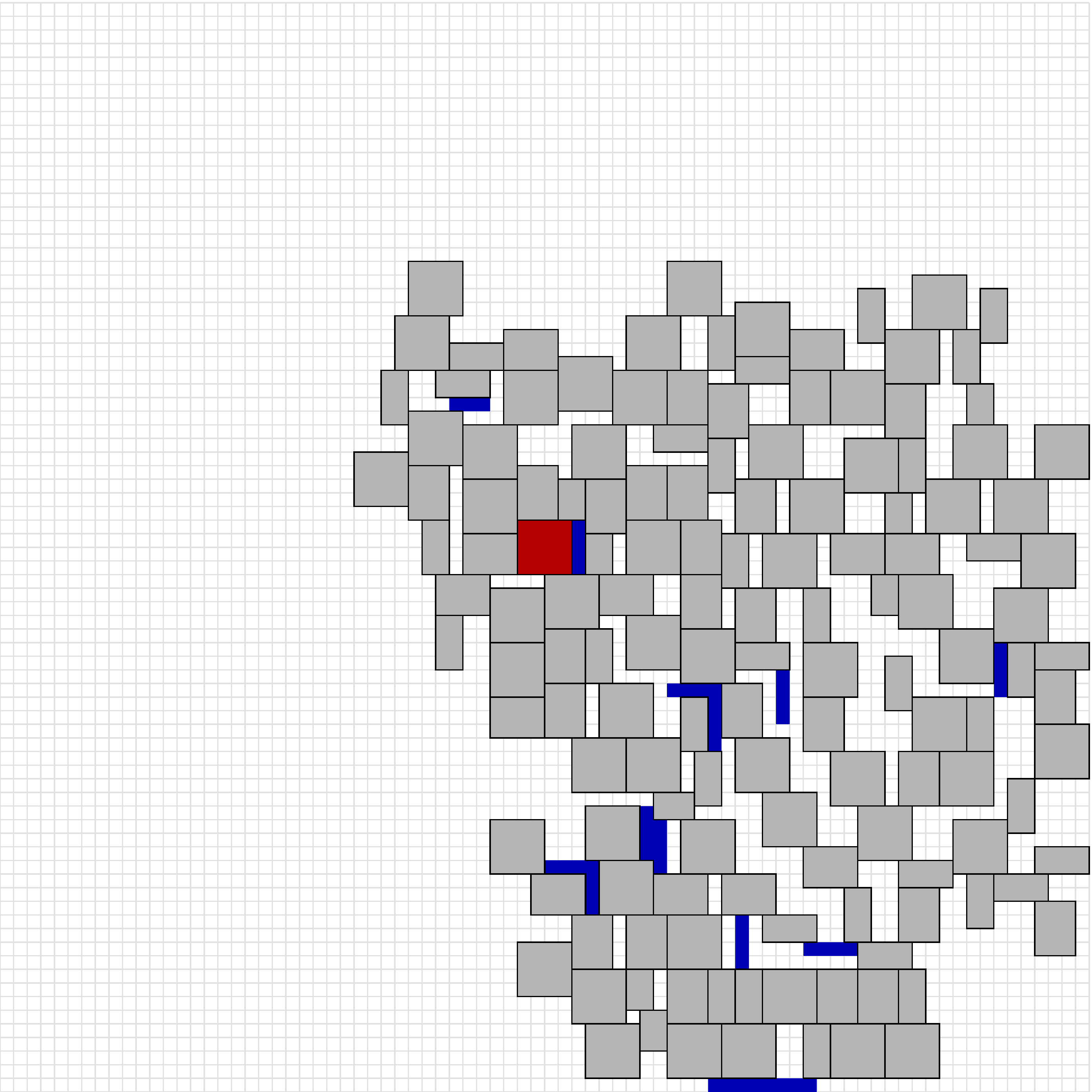}      
\includegraphics[width=0.30\textwidth]{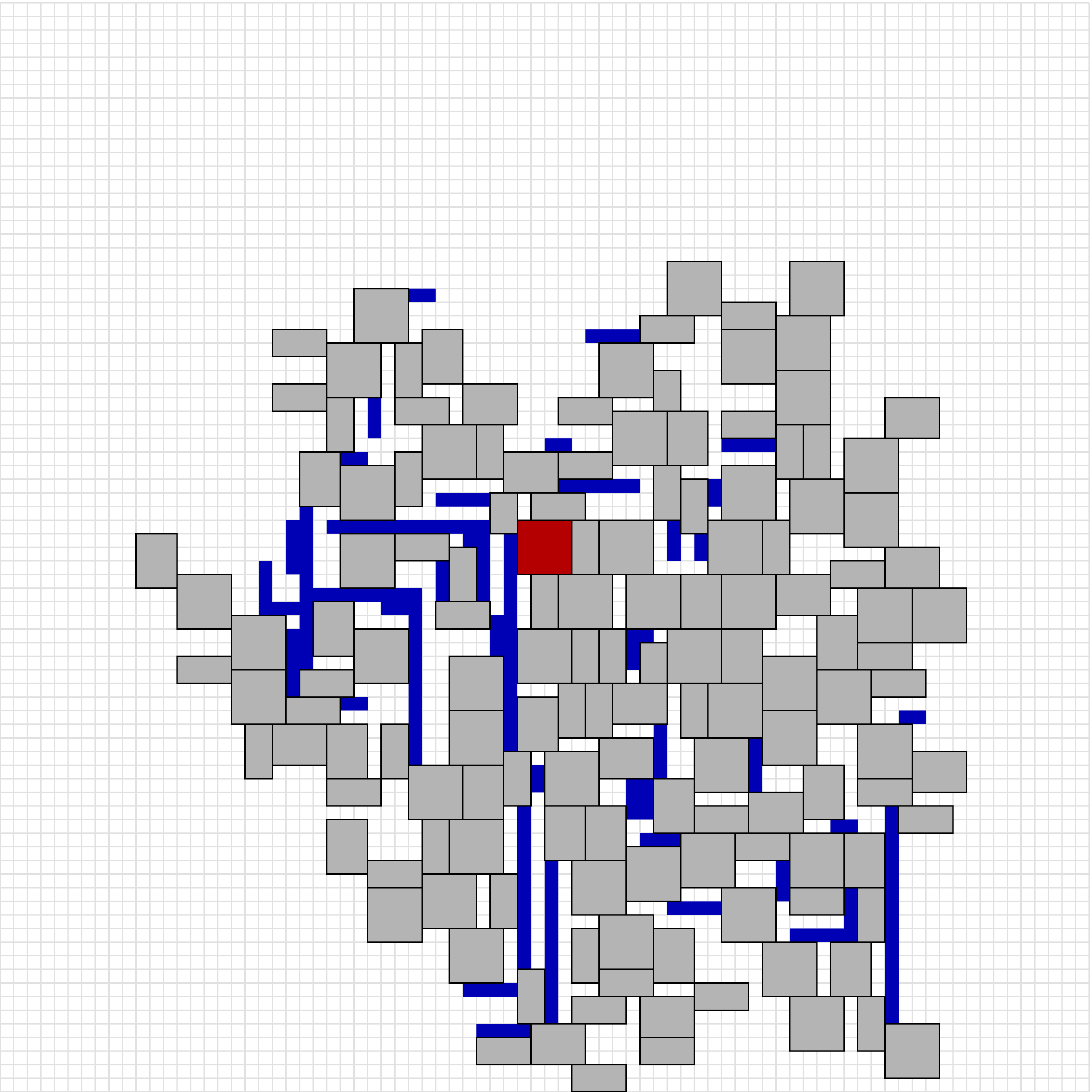}
\includegraphics[width=0.30\textwidth]{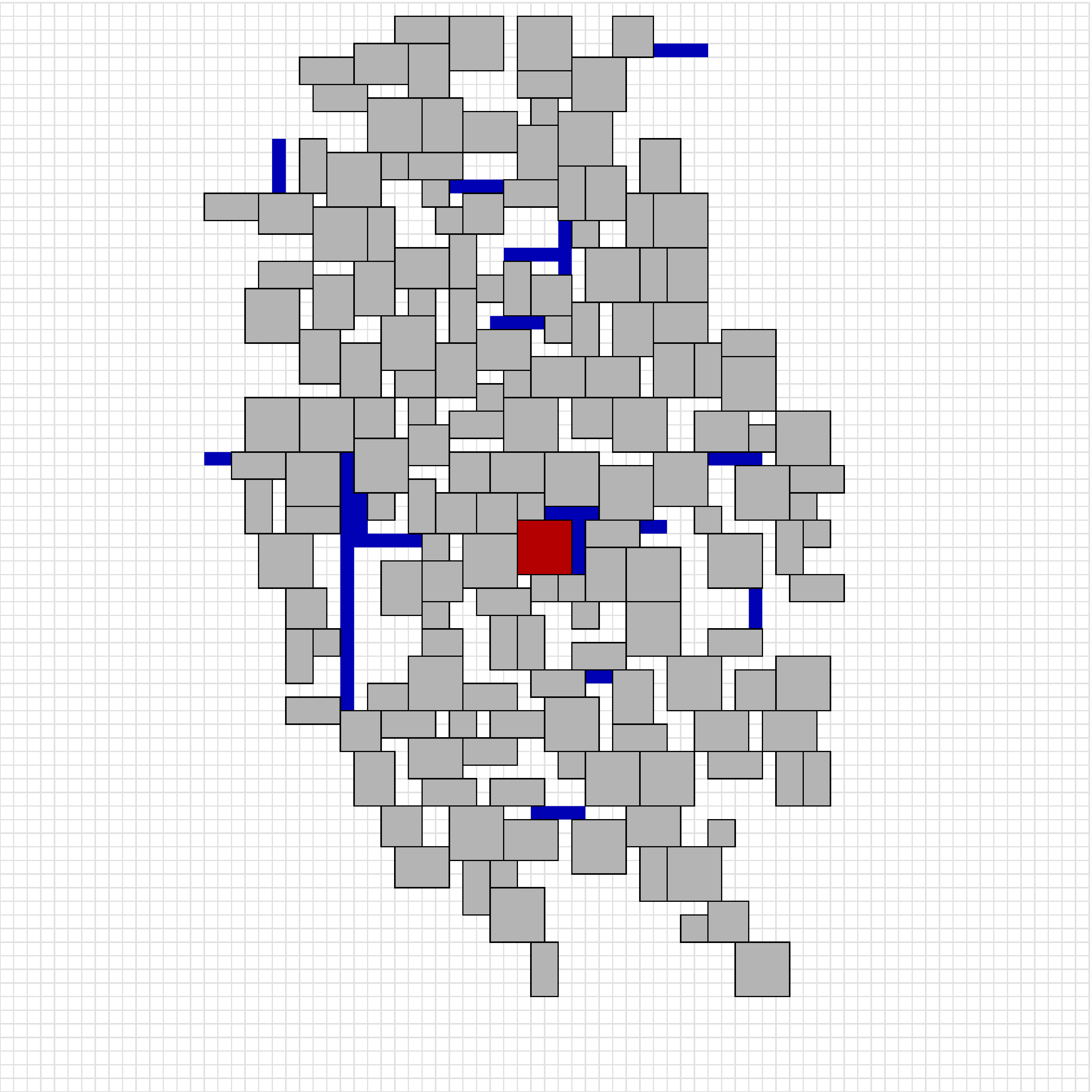}\\ 
\includegraphics[width=0.30\textwidth]{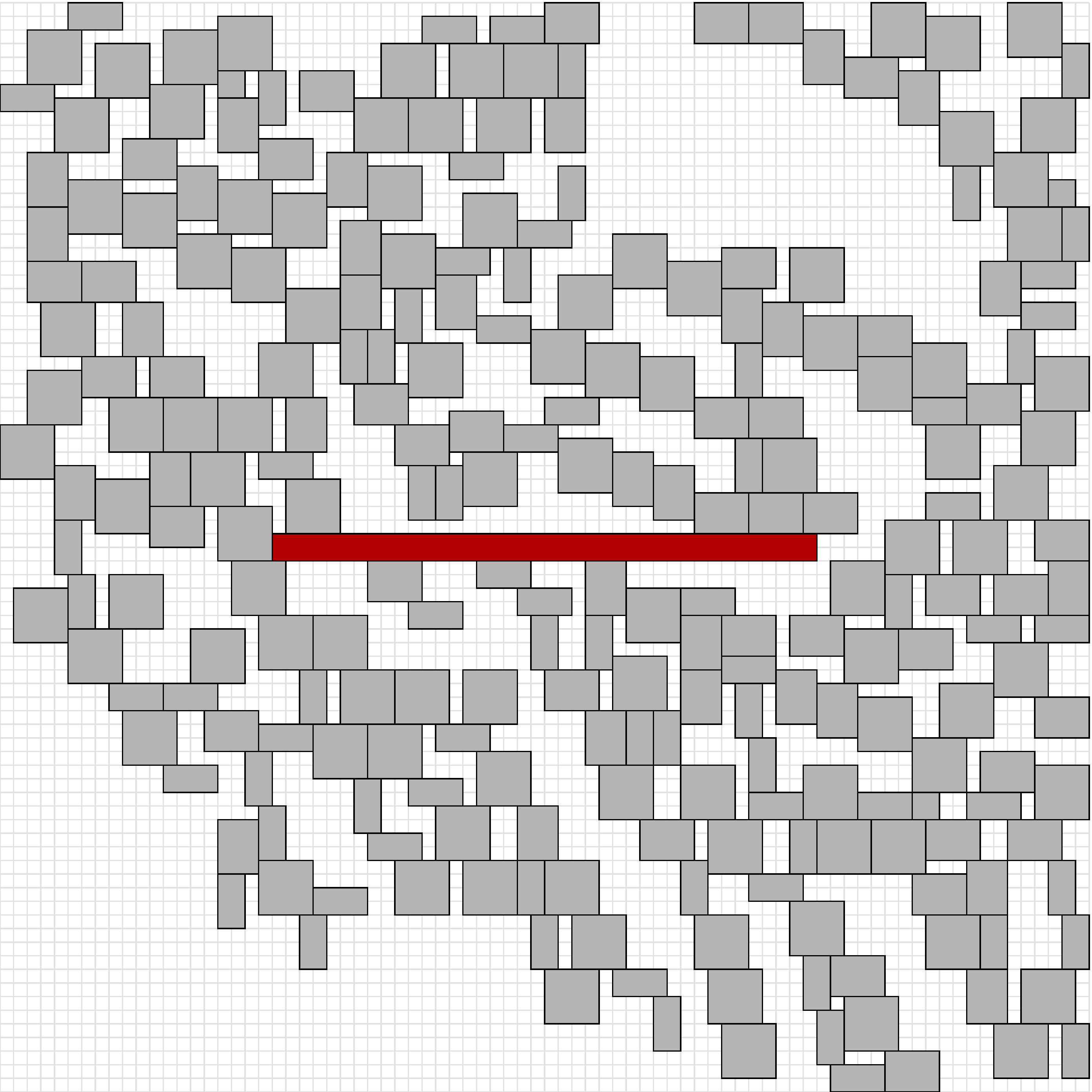}
\includegraphics[width=0.30\textwidth]{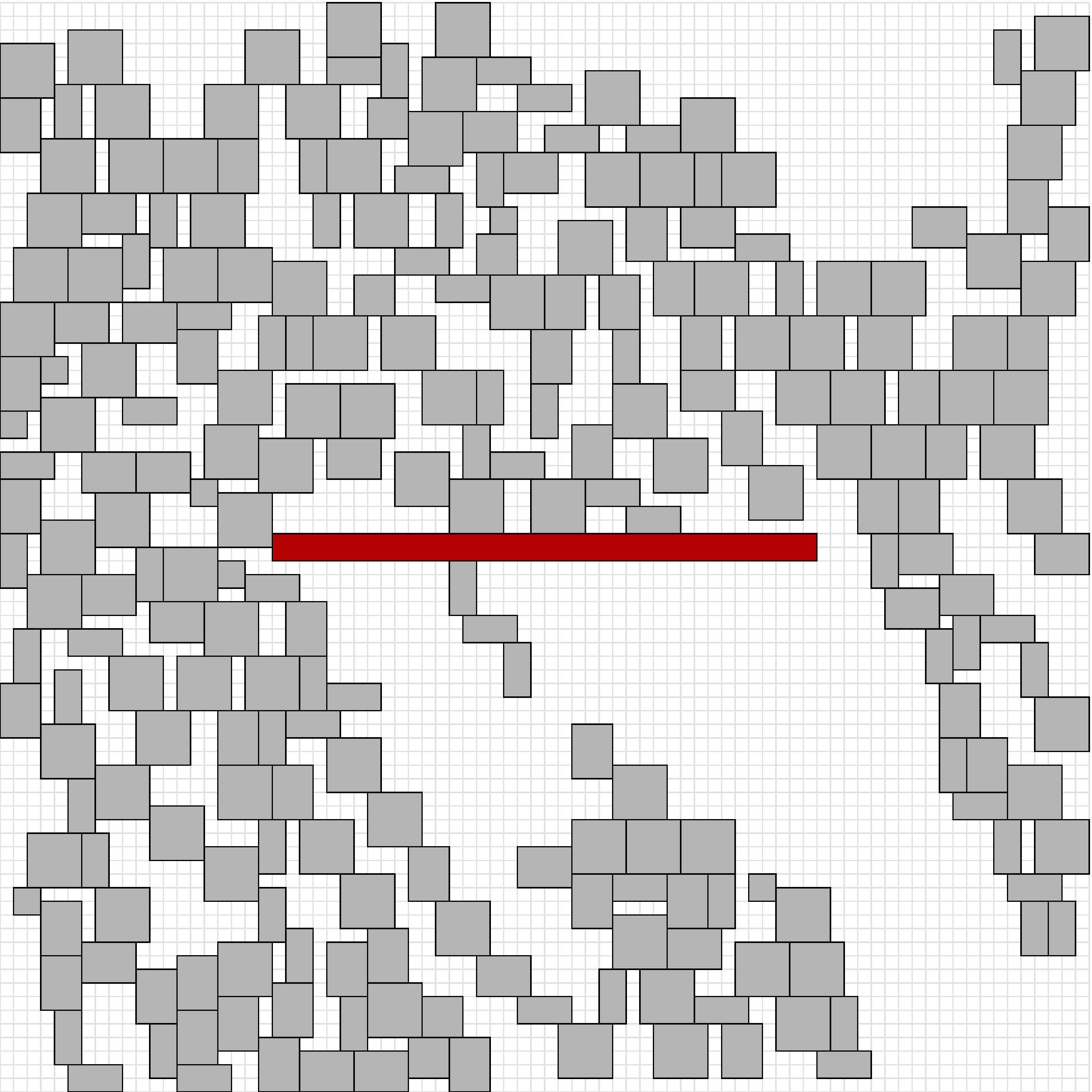}    
\includegraphics[width=0.30\textwidth]{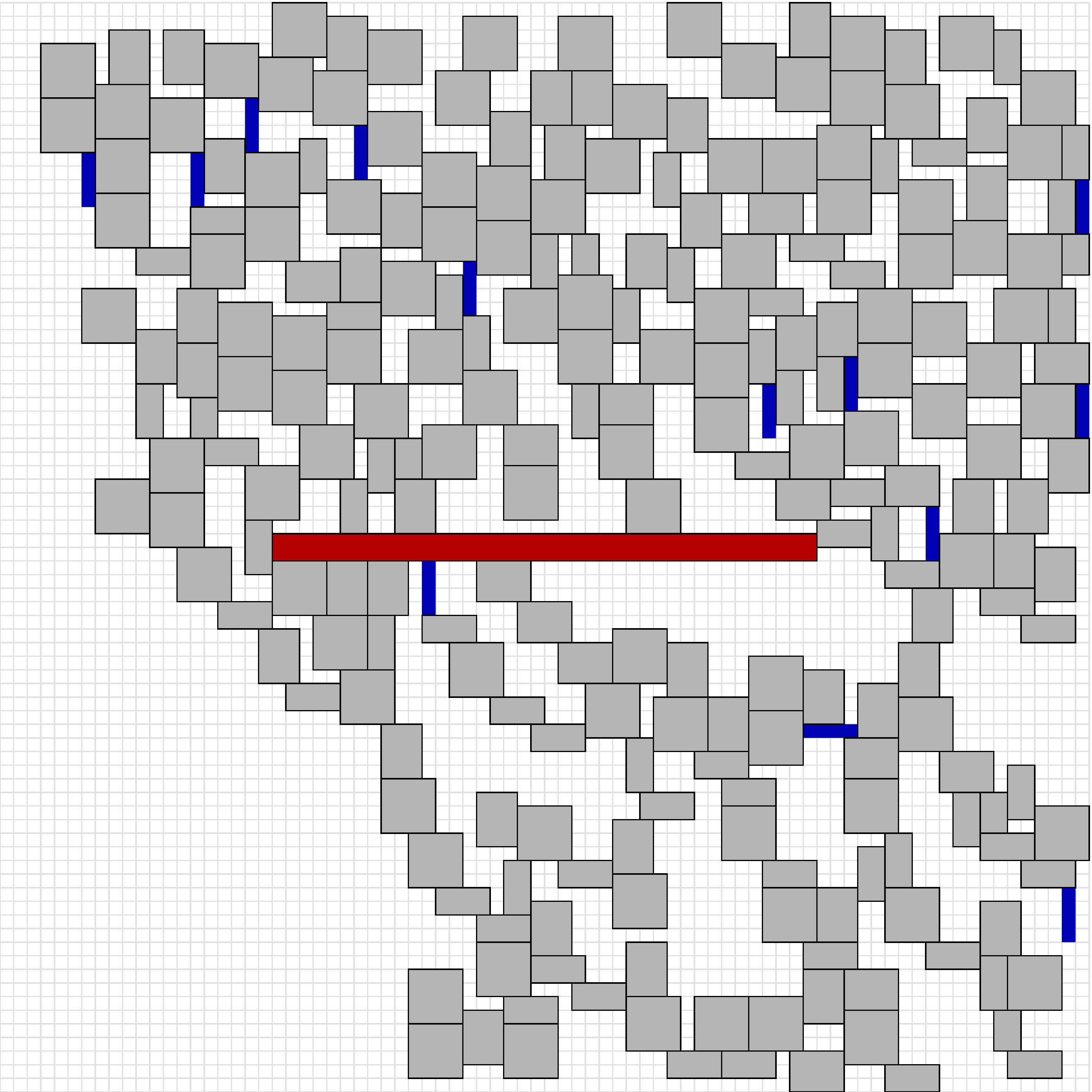}\\    
\end{tabular}
\end{center}
\caption{Three examples of evolved maps with standard and elongated
  starting rooms.  Starting rooms are shown in red.}
\label{contrast}
\end{figure*}

The final experiment examined the impact of changing the starting room.  Changing the starting room to a single, long corridor gave the algorithm far more space to expand the map and resulted in much more even placement of rooms within the grid.  A contrast demonstrating the impact of the alternate starting room appears in Figure \ref{contrast}.  The location, size, and shape of the starting room are clearly a factor that influences the
character of the map that appears.  There is no need to start with just one room -- using a system of two or three corridors could help shape the overall map and might remove the need for the recent room hack by increasing the available useful perimeter of the growing map.

\section{Conclusions and Next Steps}
\label{CNS}

This study demonstrated the ability of the do what's possible
representation to lay out very large maps.  The parameter study showed
that the system is relatively robust to the choice of parameters.  If
we think about why this system is robust to parameter choice, an
interesting hypothesis arises.

The system presented here proposes a huge stream of rooms via its SDA
and the bit-slicing decision process.  These rooms proposals are then
filtered by the GPF to yield a map.  If we tracked the number of
proposed rooms required to achieve the final result then we might see
substantially more difference arising from different parameter
choices.  The filtration of a patterned, endless stream of proposals
seems to yield an intrinsically robust system.  To make an analogy, if
students are permitted to take an on-line quiz over and over until
they like their grade, then the final results will not be proportional
to the student's effort -- less well prepared students will achieve
similar grades.  Additional study of this hypothesis is a priority for
additional research.

The runs employing additional time and space showed that the system
scales to larger problem instances, but it does not begin to test the
systems scalability.  The rooms added to a growing map are added one
at a time, and after each room is added, a connected map is available.
This means that a given evolved map builder produces a {\em sequence}
of maps.  While we stop the process at a fixed point an evaluate
fitness, we do not have to do so.  If we allow a larger arena, then we
can keep adding rooms; for some of the evolved map builders we can do
so indefinitely.  Running a map builder is quite fast, so arbitrarily
large maps are within the capabilities of the system presented here.

One concern that arose before performing this research is that,
because of the way rooms are added to existing rooms, that the map,
viewed as a combinatorial graph, would have room adjacencies that
formed a tree.  The representation completely avoided this by packing
in lots of rooms, with many adjacencies arising from fortuitous
placement rather than from deliberate action of the map builder.

Given the relatively small number of states in the SDAs, it is a little curious that more pattern is not visible in the final maps.  Motifs in the form of collections of similar groups of rooms appear, but the overall map seems haphazard.  This may be an effect caused by obstruction generated via serial addition of rooms.  As the map grows, rooms already in place prevent some proposed rooms from being placed.  The pattern of this obstruction is idiosyncratic to a particular SDA and breaks the crystalline order that might otherwise arise.

\subsection{Things that could be done}

The system presented in this study was tested on a square grid of
cells and rooms were added to a map under the influence of a very
simple fitness function.  A small number of room shapes were used and
the system constructively creates a connected map.  There are a large
number of additional directions the system can be developed in.

\begin{figure}[htb]
\centerline{\includegraphics[width=0.40\textwidth]{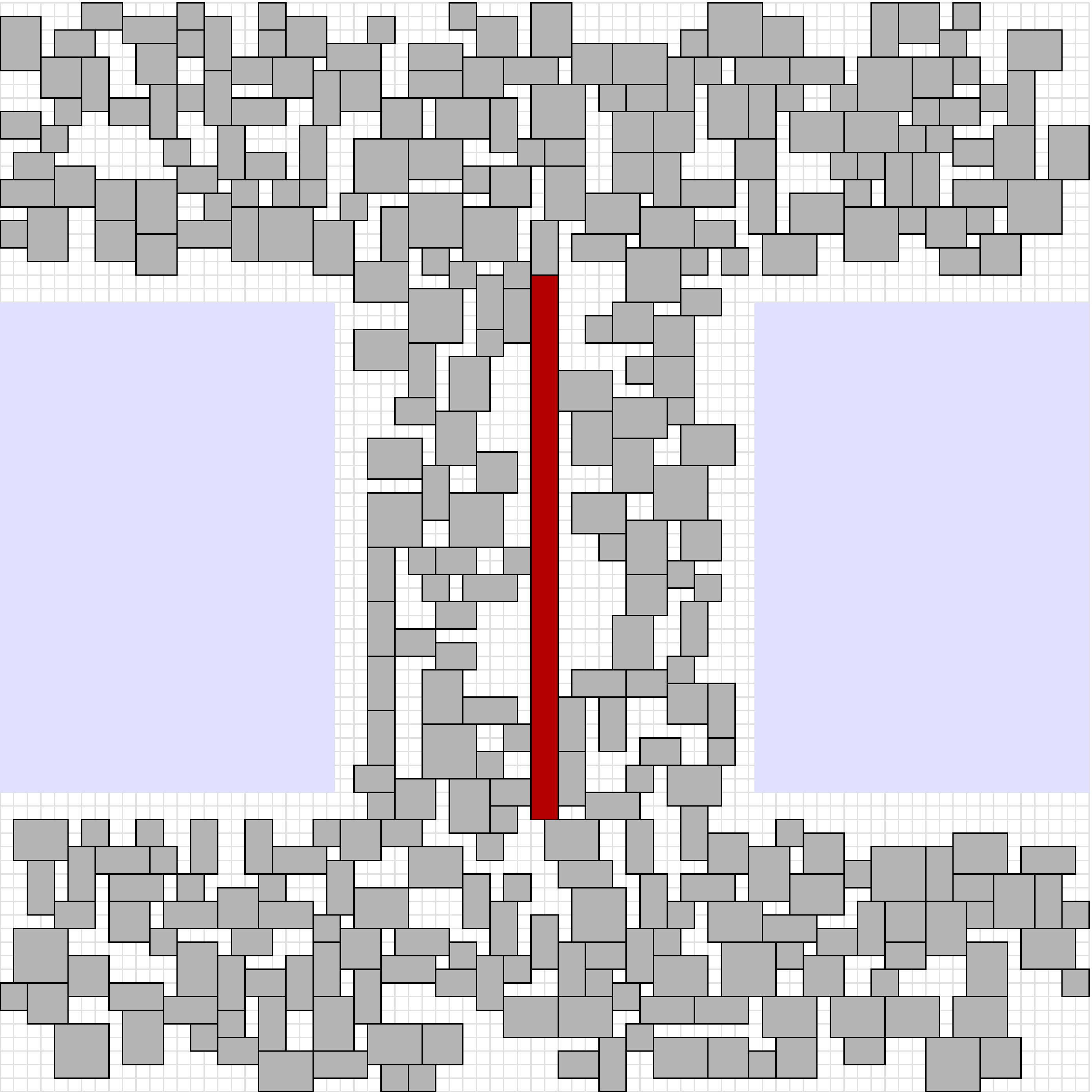}}

\vspace{0.1in}

\centerline{\includegraphics[width=0.40\textwidth]{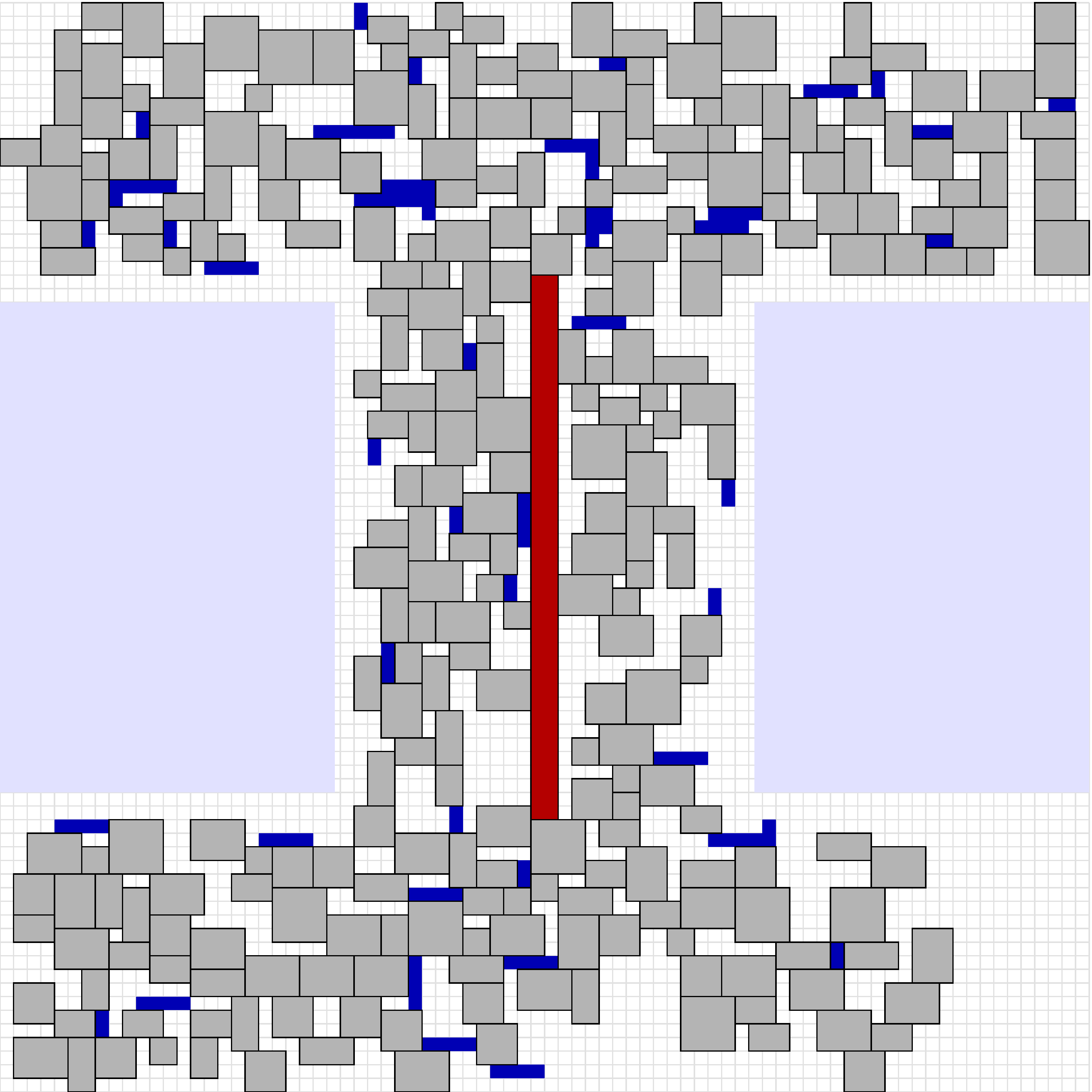}}
\caption{Shown are examples of maps shaped by placing obstructions in the bounding square of the map.  The obstructions are shaded light blue.}
\label{ChrisSays}
\end{figure}

One of the most obvious future directions follows from the fact that
the system can adapt to grow in different directions via its evolved
selection of which rooms to use as a focus for adding the next room.
If we are fitting a map into a constrained shape, with areas where
rooms cannot be built, it is almost trivial to initialize the drawing
arena to forbid rooms in some area.  A quick test of this notion was performed and appears in Figure \ref{ChrisSays}.  If we start with a high level
plan for where patches of rooms need to go, then the DWP room-layout
tool can be used to fill in details.

The fitness function used -- area of rooms placed squared divided by
the area of the bounding box -- encourages the representation to lay
out as many rooms as it can in a relatively dense form.  Many other
fitness functions are possible.  A fitness function that rewarded a
particular density of rooms would be simple to implement and would
grant the user additional control of the type of map generated, for
example.  Slightly more expensive fitness functions could use dynamic
programming to work on tactical properties of the map~\cite{Ashlock10pcg1, Ashlock16gapd}, or one could even use different AI personas to adapt maps for a given play style or play preference~\cite{Holmgard2018}.

Something left out of the current system is the placement of doors.
It seems likely that an additional developmental phase that placed
doors would be a good follow-on project. The current system insists that a new room share at least one grid's worth of wall with the room it is placed next two.  Insisting that one
or the other wall be flush, or even that the wall align, would create
a very different sort of map.  Tinkering with the rules for how rooms
are adjoined is a rich area for modifying the system, and different
room adjoinment rules can be mixed by simply adding another decision
to the SDA's decision process.

\subsection{Avoiding the need for the recent room hack}

The recent room hack avoids the problem of selecting ``used up''
rooms, those with no adjacent space, as the foci for addition of a new
room.  The fairly arbitrary choice of the most recent ten rooms is an
additional algorithmic parameter that could be explored.  There are
two other ways to address the issue that the hack addresses.  The
first is a blacklist for rooms.

The blacklist would require a critical number of $R$ rejections.  If a
room is proposed by the SDA as a place to site an additional room and
rejected $R$ times, it is removed from the list of available rooms.
This would require maintaining a list of available rooms, not a
difficult task, and would allow the system to refine and improve its
GPF with information gained during run time.  This modification of the
algorithm would probably substantially improve performance.

A second and much simpler way to avoid the need for the random room
hack is implicit in the final set of experiments with the elongated
starting room.  The system can accept multiple starting rooms as
initial conditions.  Build a sparse network of corridors as the
initial state of the system.  Then the active area where rooms can be
placed starts very large and avoids, at least for quite a while, the
congestion that plagued the system with an initial $4\times 4$ room.

\subsection{Different room shapes and required content}

In \cite{Cam11}, rooms with particular shapes were specified ahead of
time and incorporated into procedurally generated maps.  In this
study, rectangular rooms were used for ease of placement.  A small
upgrade to the code that determines if a room can be placed in a given
location would permit the DWP map builder to use a far wider variety
of rooms.

If very specific rooms were required they could be placed in a list
and one of the possibilities in the CSG's decision process would be an
attempt to place a special room.  The special room remains in the list
until placed or, possibly, until some number of instances of it have
been placed.

\subsection{SDA are not the only choice for CSGs}

In the original DWP research \cite{AshlockDWP16}, SDAs were not the
only evolvable structures used as CSGs.  Another technology, called an
{\em alternator list} was also used.  SDAs were chosen for this study
because they can generate aperiodic stings of characters while
alternator lists must eventually fall into a cyclic pattern; the very
long period of evolved alternator lists makes this a largely
irrelevant choice.

In fact, though, the maps located with SDAs in this study were a good
deal less structured that the authors had originally anticipated.
Using a relatively simple periodic stream of characters might yield
interesting maps, even a relatively short string repeated
indefinitely.  In any case, exploration of different choices of
complex string generators is a rich area for future research.

\subsection{Adaptivity}

One particular strength of this approach is that it is adaptive on two
different levels. There is a powerful analogy for this in biology - a
plant, for example, adapts through evolution and might acquire certain
growth pattern, but then it also adapts to its actual environment
during its lifetime, by growing into a certain shape, or around
obstacles. The representation discussed here has a similar split. The
CSG can evolve to encode a specific approach to the problem, i.e. what
patterns to use, or what solution to try first, while the GPF can
adapt the suggested solution to the concrete problem in
hand. Fig.\ref{ChrisSays} shows how the rooms are placed around
obstacles. In this example it would be possible to evolve the CSG for
a different set of obstacles - creating a typical pattern or built
style, and then apply this built style to different obstacle patterns.

This two stage adaptivity could be an interesting new approach for the
challenges in adaptive procedural content generation, which in
contrast to \textit{tabula rasa} PCG, needs to create content in
response to already existing content. This adaptivity to existing
content should not be confused with adaptivity to the
player~\cite{5765665}, where generated content adapts to play styles
or a player's competence level. An example for the former is the AI
Settlement Generation Challenge~\cite{salge2018generative}, where
participants have to write an algorithm that can generate a city for
an unseen Minecraft map. The solution should, among other things,
adapt to the underlying terrain. The DWP representation could be used,
by providing an appropriate complex string to road element mapping to
build a road network that would avoid inappropriate road
placement. The CSG could be evolved to produce a style of road network
that would fit certain criteria - or a curiosity driven approach could
be used to generate road networks of different styles, and those could
then be applied and conform to different terrain.

In this context it would then also make sense to soften the evaluation
of the GPF to a ``whats affordable'' representation, similar to the
approach in \cite{Ashlock16agr}. Placing a new piece of road could not
just be allowed or forbidden, but the output string could be used to
choose from a distribution of options, with better, more affordable
options having a higher likelihood to be chosen. So the algorithm
might prefer a flat road on even terrain to a step road over rugged
terrain.

\section{Summary}

In this paper we show how the ``do what's possible'' representation
can be applied to the concrete problem of procedural dungeon
generation. Compared to the range of other dungeon generators this
approach is relatively light-weight in its representation - which
reduces the search space a genetic algorithm has to go through. Yet
the representation is scalable, capable of creating dungeons of
arbitrary size, and grows dungeons iteratively in such a way that they
process could be stopped at any time with a solution. Of course, this
is traded off with various advantages and disadvantages compared to
other generators. We also discussed a range of minor modifications
that would allow one to enhance or fine tune the output for practical
application. A particularly noteworthy feature is the two-tier
adaptivity, that would allow for content generation that could be
sensitive to existing content, bearing a suitable chosen generative
possibility filter. The last property might make this approach
particularly useful to generate game content that would typically grow
organically over time in response to external stimuli, such as plants,
cities, dungeons, etc.

\bibliography{reference}

\begin{thebibliography}{10}

\bibitem{Ashlock06a}
D.~Ashlock.
\newblock {\em Evolutionary Computation for Optimization and Modeling}.
\newblock Springer, New York, 2006.

\bibitem{Ashlock15fbca}
D.~Ashlock.
\newblock Evolvable fashion-based cellular automata for generating cavern
  systems.
\newblock In {\em Proceedings of the 2015 IEEE Conference on Computatational
  Intelligence in Games}, pages 306--313, 2015.

\bibitem{Ashlock16fbca}
D.~Ashlock and L.~Bickley.
\newblock Rescalable, replayable maps generated with evolved cellular automata.
\newblock In {\em Acta Physica Polonica (B), Proceedings Supplement}, volume
  9(1), pages 13--22, 2016.

\bibitem{AshlockDWP17}
D.~Ashlock, S.~Gillis, and W.~Ashlock.
\newblock Infinite string block matching features for dna classification.
\newblock In {\em Proceedings of the 2017 IEEE Conference on Computational
  Intelligence in Bioiformatics and Computational Biology}, pages 1--8,
  Piscataway NJ, 2017. IEEE Press.

\bibitem{AshlockDWP16}
D.~Ashlock, S.~Gillis, A.~McEachern, and J.~Tsang.
\newblock The do what's possible representation.
\newblock In {\em Proceedings of the IEEE 2016 Congress on Evolutionary
  Computation}, pages 1586--1593, Piscataway NJ, 2016. IEEE Press.

\bibitem{Ashlock10pcg1}
D.~Ashlock, C.~Lee, and C.~Mc{G}uinness.
\newblock Search based procedural generation of maze like levels.
\newblock {\em IEEE Transactions on Computational Intelligence and AI in
  Games}, 3(3):260--273, 2011.

\bibitem{Ashlock10pcg2}
D.~Ashlock, C.~Lee, and C.~Mc{G}uinness.
\newblock Simultaneous dual level creation for games.
\newblock {\em Computational Intelligence Magazine}, 2(6):26--37, 2011.

\bibitem{Ashlock11pcg1}
D.~Ashlock and C.~McGuinness.
\newblock Decomposing the level generation problem with tiles.
\newblock In {\em Proceedings of CEC 2011}, pages 849--856, 2011.

\bibitem{Ashlock13pcg}
D.~Ashlock and C.~McGuinness.
\newblock Landscape automata for search based procedural content generation.
\newblock In {\em Proceedings of IEEE CIG 2013}, pages 9--16, 2013.

\bibitem{Ashlock16gapd}
D.~Ashlock and C.~McGuinness.
\newblock Graph-based search for game design.
\newblock {\em Game and Puzzle Design}, 2(2):68--75, 2016.

\bibitem{Ashlock16agr}
D.~Ashlock and J.~Montgomery.
\newblock An adaptive generative representation for evolutionary computation.
\newblock In {\em Proceedings of the IEEE 2016 Congress on Evolutionary
  Computation}, pages 1578--1585, Piscataway NJ, 2016. IEEE Press.

\bibitem{Holmgard2018}
C.~{Holmgard}, M.~C. {Green}, A.~{Liapis}, and J.~{Togelius}.
\newblock Automated playtesting with procedural personas with evolved
  heuristics.
\newblock {\em IEEE Transactions on Games}, pages 1--1, 2018.

\bibitem{Johnson10}
Lawrence Johnson, Georgios~N. Yannakakis, and Julian Togelius.
\newblock Cellular automata for real-time generation of infinite cave levels.
\newblock In {\em Proceedings of the 2010 Workshop on Procedural Content
  Generation in Games}, PCGames '10, pages 10:1--10:4, New York, NY, USA, 2010.
  ACM.

\bibitem{5765665}
R.~{Lopes} and R.~{Bidarra}.
\newblock Adaptivity challenges in games and simulations: A survey.
\newblock {\em IEEE Transactions on Computational Intelligence and AI in
  Games}, 3(2):85--99, June 2011.

\bibitem{McCormack2005GenerativeDA}
Jon McCormack and Alan Dorin.
\newblock Generative design: a paradigm for design research.
\newblock In {\em Proceedings of Futureground}, 2005.

\bibitem{Cam11}
C.~McGuinness and D.~Ashlock.
\newblock Incorporating required structure into tiles.
\newblock In {\em Proceedings of IEEE CIG 2011}, pages 16--23, Piscataway NJ,
  2011. IEEE Press.

\bibitem{salge2018generative}
Christoph Salge, Michael~Cerny Green, Rodgrigo Canaan, and Julian Togelius.
\newblock Generative design in minecraft (gdmc): settlement generation
  competition.
\newblock In {\em Proceedings of the 13th International Conference on the
  Foundations of Digital Games}, page~49. ACM, 2018.

\bibitem{smith2014logical}
Anthony~J Smith and Joanna~J Bryson.
\newblock A logical approach to building dungeons: Answer set programming for
  hierarchical procedural content generation in roguelike games.
\newblock In {\em Proceedings of the 50th Anniversary Convention of the AISB},
  2014.

\bibitem{Sorenson10a}
N.~Sorenson and P.~Pasquier.
\newblock Towards a generic framework for automated video game level creation.
\newblock In {\em Proceedings of the European Conference on Applications of
  Evolutionary Computation (EvoApplications)}, volume 6024, pages 130--139.
  Springer LNCS, 2010.

\bibitem{Julian10a}
Julian Togelius, Mike Preuss, and Georgios~N. Yannakakis.
\newblock Towards multiobjective procedural map generation.
\newblock In {\em PCGames '10: Proceedings of the 2010 Workshop on Procedural
  Content Generation in Games}, pages 1--8, New York, NY, USA, 2010. ACM.

\bibitem{TogeliusSBPCG}
Julian Togelius, Georgios Yannakakis, Kenneth Stanley, and Cameron Browne.
\newblock Search-based procedural content generation.
\newblock In {\em Applications of Evolutionary Computation}, volume 6024 of
  {\em Lecture Notes in Computer Science}, pages 141--150. Springer Berlin /
  Heidelberg, 2010.

\bibitem{Wilson01}
J.~H. van Lint and R.~M. Wilson.
\newblock {\em A Course in Combinatorics, second edition}.
\newblock Cambridge University Press, New York, NY, 2001.

\end{thebibliography}

\end{document}